# Simulated Annealing for Emotional Dialogue Systems

Chengzhang Dong, Chenyang Huang, Osmar Zaïane, Lili Mou
Dept. Computing Science & Alberta Machine Intelligence Institute (Amii)
University of Alberta, Edmonton, AB, Canada
{chengzha,chuang8,zaiane}@ualberta.ca,doublepower.mou@gmail.com

## ABSTRACT

Explicitly modeling emotions in dialogue generation has important applications, such as building empathetic personal companions. In this study, we consider the task of expressing a specific emotion for dialogue generation. Previous approaches take the emotion as an input signal, which may be ignored during inference. We instead propose a search-based emotional dialogue system by simulated annealing (SA). Specifically, we first define a scoring function that combines contextual coherence and emotional correctness. Then, SA iteratively edits a general response and searches for a sentence with a higher score, enforcing the presence of the desired emotion. We evaluate our system on the NLPCC2017 dataset. Our proposed method shows 12% improvements in emotion accuracy compared with the previous state-of-the-art method, without hurting the generation quality (measured by BLEU).[1]

## CCS CONCEPTS

• **Computing methodologies** → **Natural language generation**.

## KEYWORDS

Dialogue generation, Simulated annealing, Emotion analysis

## 1 INTRODUCTION

Building dialogue systems is a challenging task because of the difficulty for computers to imitate human speech behaviors and emotions. Recently, the sequence-to-sequence (Seq2Seq) model is a common approach to dialogue systems, but Seq2Seq tends to generate short and dull responses, such as "thank you" and "me too" [8, 15].

Explicitly modeling emotions becomes an important research direction to alleviate the dull responses given by a plain Seq2Seq [1, 13, 27]. Further, an emotional dialogue system can serve as an empathetic companion system for humans. For example, an old and lonely person may feel down and anxious at home, and an emotional chatbot can bring optimistic and positive attitudes.

Our study considers the setting of generating a response, given an input utterance and a target emotion. In practice, the target emotion may be predicted by another machine learning model. Nevertheless, we focus on the generation part and assume the target emotion is given, following most previous settings [6, 20, 21, 28].

In previous work, Song et al. [21] propose an attention mechanism to emotion lexicons; Zhou et al. [28] build a model, called the emotional chatting machine (ECM), that uses both emotion lexicon attention and the target emotion embedding to enhance a Seq2Seq model. Such training signals, however, are indirect, as the neural model may ignore these input emotion embeddings and lexicons during the training process. Shen et al. [20] propose a curriculum dual learning approach based on ECM and reinforcement learning (RL), involving both content-based and emotion-based rewards. However, RL is difficult to train due to the large search space.

In this paper, we propose a search-based emotional dialogue system by simulated annealing (SA). SA is a stochastic search algorithm towards an objective function. We design the search objective as the multiplication of Seq2Seq's generation probability and an emotion classifier's probability. Our work is inspired by the recent development of SA for unsupervised text generation [12], but we adapt it to the Seq2Seq setting for emotional dialogue generation. SA starts from Seq2Seq's output, but searches towards the objective by local editing. For each search step, our system first proposes a candidate sentence by an editing operation, namely, word insertion, replacement, and deletion. The proposed candidate can either be accepted or rejected based on its score given by the search objective. If accepted, it is used for the next iteration. In SA, a temperature controls the acceptance possibility: at the beginning, SA tends to accept a low-scored sentence, and as the search proceeds, the temperature is cooled down for better convergence. By explicitly performing the search during inference, our model can better enforce the presence of the desired emotion. Our edit-based local search is also more effective than RL, which learns to generate an utterance word-by-word in a trial-and-error fashion.

We evaluated our approach on the NLPCC2017 dataset. Experimental results show that our model equipped with diverse beam search achieves an improvement of 12 percentage points in emotion accuracy, compared with previous state-of-the-art methods. Meanwhile, our SA search does not hurt the general dialogue quality measured by BLEU scores.

## 2 APPROACH

We formulate the task of emotional dialogue generation as follows. Let $\mathbf{x} = (x_1, \cdots, x_S)$ be an input utterance and $e$ be a target emotion (e.g., "happy" or "sad"). We would like to generate an utterance $\mathbf{y} = (y_1, \cdots, y_T)$ being the response of $\mathbf{x}$ and having the emotion $e$. Here, $S$ and $T$ are the lengths of input and output, respectively. As mentioned in §1, we assume the target emotion is always given during training and inference, and in practice, the emotion may be either pre-defined or predicted.

In the rest of this section, we first introduce a general Seq2Seq dialogue system (§2.1). Then, we design a search objective considering both Seq2Seq probability and desired emotion probability (§2.2). Finally, we introduce the simulated annealing search that maximizes the objective for emotional dialogue systems (§2.3). Figure 1 shows an overview of our approach.

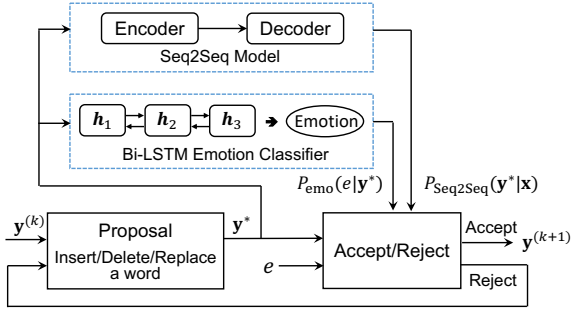

Figure 1: Overview of our simulated annealing approach to emotional dialogue generation.

## 2.1 Seq2Seq For Dialogue Generation

The Seq2Seq model [22] serves as the backbone of our dialogue generation system. It has an encoder that processes the input $\mathbf{x}$ and a decoder that predicts each word in $\mathbf{y}$ in an autoregressive fashion. We use the long short-term memory (LSTM) model as both the encoder and decoder with an attention mechanism [23].

The encoder first represents the discrete tokens in $\mathbf{x}$ as vectors by an embedding layer, denoted as $\boldsymbol{x}_i = \text{emb}(x_i)$. Then, the encoder applies LSTM on these embeddings, and all the LSTM hidden states will be used in the decoder by attention. The encoder can be denoted by

$$\mathbf{H}_{\text{Enc}} = \text{Encoder}(\boldsymbol{x}_1, \cdots, \boldsymbol{x}_S) \quad (1)$$

The decoder predicts every word $\mathbf{y}_i$ conditioned on the previously predicted words $\mathbf{y}_{<i}$, as well as the input by an attention to $\mathbf{H}_{\text{Enc}}$. We formulate the process as follows.

$$\boldsymbol{h}_i = \text{Decoder}(\mathbf{H}_{\text{Enc}}, \boldsymbol{y}_1, \cdots, \boldsymbol{y}_{i-1}) \quad (2)$$
$$P_{\text{Seq2Seq}}(y_i|\mathbf{x}, \mathbf{y}_{<i}) = \text{softmax}(\mathbf{W}_o \boldsymbol{h}_i) \quad (3)$$

where $\mathbf{W}_o$ is a weight matrix and $\boldsymbol{y}_j = \text{emb}(y_j)$ is the embedding of a previously generated word for $j = 1, \cdots, i-1$.

Seq2Seq learns a conditional language model, which maximizes $P_{\text{seq2seq}}(\mathbf{y}|\mathbf{x})$ during training. For inference, it predicts the response $\mathbf{y}$ by maximizing the conditional probability either greedily or with beam search.

Note that a plain Seq2Seq does not model the emotion $e$ while generating $\mathbf{y}$, as $e$ is not involved in the model. Previous work [20, 21, 28] modifies the Seq2Seq model by takeing $e$ as an additional input, and learns the probability of $p(\mathbf{y}|\mathbf{x}, e)$ in an end-to-end fashion. However, because such an approach only relies on end-to-end training for emotion treatment, it may ignore the input $e$, thus still lacking explicit control over the emotion.

## 2.2 Search Objective

We propose to explicitly incorporate the emotion into the decoding objective for emotional dialogue systems. We define a scoring function $f(\mathbf{y}; \mathbf{x}, e)$ to evaluate the quality of a candidate response $\mathbf{y}$ given an input utterance $\mathbf{x}$ and the desired emotion $e$, considering both contextual coherence (based on the Seq2Seq in §2.1) and emotion correctness (based on an emotion classifier). The scoring function will be used as the search objective for decoding.

Specifically, we train a bidirectional LSTM for encoding an utterance $\mathbf{u}$. Then, a Transformer-like attention mechanism [23] computes the weighted average of Bi-LSTM hidden states as $\boldsymbol{h}_{\text{emo}}$ for softmax emotion classification:

$$P_{\text{emo}}(e|\mathbf{u}) = \text{softmax}(\mathbf{W}_e \boldsymbol{h}_{\text{emo}}) \quad (4)$$

where $\mathbf{W}_e$ is a weight matrix. The classifier is trained with an emotion-labeled corpus, and in our experiment, we used the same dataset as [20, 21, 28].

We apply $P_{\text{emo}}$ to evaluate the emotion of a candidate response $\mathbf{y}$. If the target emotion $e$ has the largest value $P_{\text{emo}}(\cdot|\mathbf{y})$, then the response indeed expresses the desired emotion.

Finally, we combine the emotion correctness with the contextual coherence to measure the quality of a response, given by

$$f(\mathbf{y}; \mathbf{x}, e) = P_{\text{Seq2Seq}}(\mathbf{y}|\mathbf{x}) \cdot P_{\text{emo}}(e|\mathbf{y})^\alpha \quad (5)$$

where $\alpha$ is a hyperparameter adjusting the weight between the two terms.

Our search objective consists of two parts: a Seq2Seq model to ensure the response $\mathbf{y}$ is contextually coherent to the input utterance $\mathbf{x}$, and an emotion classifier to ensure the specified emotion $e$ is indeed expressed in the response $\mathbf{y}$. Different from previous work [20, 21, 28], the scoring function (5) enables our approach to actively search for a proper and emotional response.

## 2.3 Search by Simulated Annealing (SA)

We use the simulated annealing (SA) algorithm to search for a desired utterance. SA starts from a general dialogue response $\mathbf{y}^{(0)} = \arg\max P_{\text{Seq2Seq}}(\mathbf{y}|\mathbf{x})$, obtained by standard beam search (BS) or diverse beam search (DBS) on the trained Seq2Seq model. Then, SA maximizes the scoring function (5) by iteratively editing the candidate response.

Let $\mathbf{y}^{(k)} = (y_1^{(k)}, \cdots, y_T^{(k)})$ be the sentence of the $k$th search step. Our model randomly chooses an editing position $t$ from $1, \cdots, T$, and an edit operation, namely, word replacement, insertion, or deletion. For deletion, we simply remove the chosen word. For insertion and replacement, a candidate word $w^*$ needs to be proposed, which is given by the posterior distribution (known as a Gibbs step):

$$P(w^*|\cdot) = \frac{1}{Z} f(\mathbf{y}^*; \mathbf{x}; e), \quad Z = \sum_{w' \in \mathcal{V}} f(\mathbf{y}'; \mathbf{x}; e) \quad (6)$$

where $\mathcal{V}$ is the vocabulary, and $\mathbf{y}^*$ is the resulting utterance when the $t$th word is replaced to $w^*$.

If the newly proposed response $\mathbf{y}^*$ has a higher score than $\mathbf{y}^{(k)}$, we accept $\mathbf{y}^*$ for further editing, i.e., $\mathbf{y}^{(k+1)} = \mathbf{y}^*$. Otherwise, we tend to reject $\mathbf{y}^*$, i.e., $\mathbf{y}^{(k+1)} = \mathbf{y}^{(k)}$. However, we may still accept a worse candidate with a small probability, given by

$$P(\text{accept }|\mathbf{y}^*, \mathbf{y}^{(k)}, e, \tau) = \min\{1, \exp(\frac{f(\mathbf{y}^*; \mathbf{x}, e) - f(\mathbf{y}; \mathbf{x}, e)}{\tau})\}, \quad (7)$$

where $\tau$ is an annealing temperature, scheduled by $\tau = \max\{0, \tau_{\text{init}} - C \cdot t\}$ for some initial temperature $\tau_{\text{init}}$ and some annealing rate $C$. In other words, SA starts with a high temperature $\tau$ at the beginning, being less greedy and more exploratory of the search space. Then, the temperature gradually cools down for better convergence. Liu et al. [12] empirically showed the effect of $\tau_{\text{init}}$ and $C$, and we adopted these hyperparameters from their paper.

|  | Models | BLEU Scores | | Diversity | | Embedding-Based Metrics | | | | Emotion |
|---|---|---|---|---|---|---|---|---|---|---|
|  |  | BLEU-1 | BLEU-2 | Dist-1 | Dist-2 | Average | Greedy | Extreme | Coherence | accuracy |
| Previous | Seq2Seq | 4.24 | 0.73 | 0.035 | 0.119 | 0.497 | 0.328 | 0.352 | 0.582 | 0.244 |
|  | EmoEmb | 7.22 | 1.64 | 0.040 | 0.133 | 0.532 | 0.356 | 0.381 | 0.594 | 0.693 |
|  | EmoDS | 9.76 | 2.82 | 0.050 | 0.174 | 0.623 | 0.403 | 0.427 | 0.603 | 0.746 |
|  | ECM | 10.23 | 3.32 | 0.052 | 0.177 | 0.625 | 0.405 | 0.433 | 0.607 | 0.753 |
|  | CDL | 12.54 | 3.70 | **0.065** | 0.221 | 0.642 | 0.438 | 0.457 | 0.635 | 0.823 |
| Ours | Seq2Seq BS | 10.78 | 3.11 | 0.058 | 0.215 | 0.765 | 0.543 | 0.594 | 0.690 | 0.253 |
|  | Seq2Seq BS + SA | 13.90 | 4.03 | 0.051 | **0.276** | 0.782 | **0.569** | 0.610 | 0.701 | 0.928 |
|  | Seq2Seq DBS | 12.14 | 3.89 | 0.061 | 0.209 | 0.768 | 0.545 | 0.601 | 0.699 | 0.264 |
|  | Seq2Seq DBS + SA | **14.26** | **4.12** | 0.053 | 0.239 | **0.786** | 0.556 | **0.611** | **0.703** | **0.942** |

Table 1: Results of baselines and our models, where Seq2Seq is a standard LSTM model with attention. BS and DBS refer to beam search and diverse beam search. The performance of previous models are quoted from [20]. Embedding-based metrics are grayed out for previous methods, because the embeddings used may be different; thus the metrics are not directly comparable.

The process is repeated iteratively, and the best-scored candidate is taken as the emotional dialogue response. In this way, our SA utilizes a general Seq2Seq for dialogue coherence, but performs local edits to revise the candidate response for the desired emotion.

## 3 EXPERIMENTS

### 3.1 Setup

**Datasets.** We follow [20] and use the NLPCC 2017 Emotional Conversation Generation Challenge dataset [28]. It contains more than 1 million post–response sentence pairs, each sentence labeled by an emotion: happy, angry, disgust, sad, like, or neutral. Also, we follow the split in [20] with 1,105,487, 11,720, and 2,000 samples for training, validation, and test, respectively.

To train the emotion classifier, we use NLPCC2013 and NLPCC2014 datasets, and obtain 64.7% emotion accuracy with our Bi-LSTM. Our setting follows prior work, and we achieve comparable results to previously reported numbers [20, 21, 28].

**Hyperparameters.** For simulated annealing, the initial temperature $\tau_{init}$ is set to 0.015, where the decay rate $C$ is 0.03 [12]. We set the maximum number of SA iterations to 50, which is generally enough for editing in our experiment. The emotion weight $\alpha$ is set to 8 by parameter tuning (shown in §3.2). To train the Seq2Seq model, we use the Adam optimizer, and set the learning rate to 5e-4. Following [21], we apply pre-trained Chinese word embeddings[2] for better semantic representation. For inference of the base Seq2Seq model, we set the beam size to 20 for both standard beam search and diverse beam search. For diverse beam search, we follow the suggested settings in [8]: the number of groups $G$ is set to 20 (same as the beam size); the diversity strength $\lambda$ is set to 0.5.

**Competing Models.** We compare the proposed method with the following baselines: (1) **Seq2Seq**: A Seq2Seq model with an attention mechanism [23]. It is also the base architecture used in all competing models. (2) **EmoEmb**: A Seq2Seq model using the embedding of a specified emotion as additional input to every decoding step [3, 9]. (3) **ECM**: A Seq2Seq model enhanced with the target emotion embedding and the emotion lexicon attention [28]. (4) **EmoDS**: An emotional dialogue system using a pointer network that allows the model to copy from a dictionary of emotional words [21]. (5) **CDL**: A variant of ECM applying curriculum dual learning and RL for emotional dialogue generation [20].

For our model, we use beam search (BS) and diverse beam search (DBS) to obtain the initial candidate for SA editing.

**Metrics.** Following [20, 21], we use the following evaluation metrics: (1) BLEU scores [16] on the scale of 0–100, measuring the general quality of a response in comparison with the reference. (2) Dist-1 and Dist-2 measuring the ratio of distinct unigrams and bigrams [8]. (3) Embedding-based scores calculated in various ways: Average, Greedy, Extreme [11], and Coherence [25]. (4) Emotion accuracy given by an emotion classifier. This is the main metric for evaluating if the generated response indeed has the target emotion.

### 3.2 Results and Analysis

**Overall performance.** Table 1 presents the overall results for emotional dialogue generation. As seen, the Seq2Seq model may not generate the response with the desired emotion. With SA, emotion accuracy is improved by a large degree of nearly 70 percentage points, verifying that SA can successfully perform editing and make the response emotional. Moreover, SA does not hurt, and in fact, improves BLEU scores. This is because our search objective involves not only the emotion probability $P_{emo}$, but also the general plausibility of a dialogue response, given by $P_{Seq2Seq}$. Actually, emotion modeling complies with the general evaluation criteria of dialogue systems (shown by BLEU).[3]

Notably, both our BS+SA and DBS+SA outperform all the previous models in terms of BLEU scores as well as the emotion accuracy. This shows our SA editing is indeed an effective approach for emotional dialogue generation.

**Analyzing the Search Objective.** We analyze the effect of the two factors in our search objective. Specially, we vary the relative weight $\alpha$ of $P_{emo}$, and see how the BLEU score and emotion accuracy would respond in Figure 2. As the emotion weight increases, the emotion accuracy increases monotonically. This is not surprising because the search objective emphasizes the emotion probability more. The BLEU-2 score also increases drastically when the emotion weight is small. The trend is consistent with Table 1 that modeling emotion (to a reasonable extent) improves the general quality of a dialogue system. However, the BLEU decreases if the emotion weight is too large, as the search objective over-emphasizes the emotion part and neglects the general dialogue quality. Based on our analysis, we set the emotion weight to 8.

---
[2]https://github.com/Embedding/Chinese-Word-Vectors

[3]We observe DBS may not improve Dist-1 and Dist-2 metrics. This is understandable because DBS encourages the diversity in the beam for a given input, whereas Dist-1 and Dist-2 evaluate the output diversity over the population of the entire test set.

| # | Input Utterance w/ Translate | Model | Target Emotion | Response | Translated Response |
|---|---|---|---|---|---|
| 1 | 可怜的孩子。<br>What a poor child. | Seq2seq<br>Ours | Anger | 你是谁啊?<br>你谁啊你? | Who are you?<br>Who the hell are you? |
| 2 | 生日快乐～<br>Happy birthday～ | Seq2seq<br>Ours | Like | 谢谢啦～<br>谢谢你啊!! 爱你～～ | Thank you~<br>Thank you!! Love you~~ |
| 3 | 可惜他没有，所以作品就是新闻联播<br>Unfortunately, he didn't, so his work is just *News Broadcast*. | Seq2seq<br>Ours | Disgust | 你说的是什么???<br>那新闻联播是很不好 | What are you talking about???<br>That *News Broadcast* is not good. |
| 4 | 当然～约起来～哈哈。<br>Sure~ let's hang out~haha. | Seq2seq<br>Ours | Happiness | 好的，我也想去<br>哈哈，好的，我也想 | Okay, I want to go too.<br>Haha, okay, I would also like to. |
| 5 | 有点心痛<br>A little bit upset | Seq2seq<br>Ours | Sadness | 你说的是什么???<br>是很心痛的啊。。 | What are you talking about???<br>Yes, it is a little bit upset .. |

Table 2: Examples of generated responses.

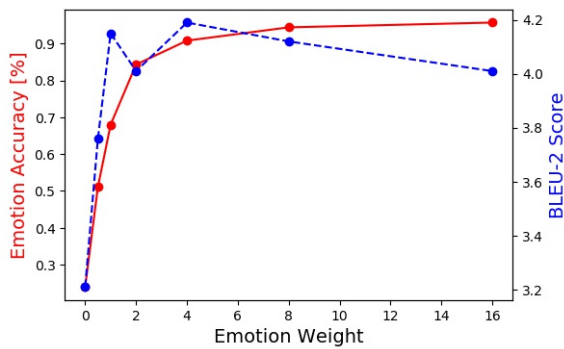

Figure 2: The emotion accuracy (solid line, left $y$-axis) and BLEU-2 (dashed line, right $y$-axis) versus the emotion weight.

**Case Study.** In Table 2, we show examples of our SA output in comparison with Seq2Seq. As seen, SA is able to insert words of the target emotion, e.g., "*not good*" for Disgust in Example 3 and "*Haha*" for Happiness in Example 4. Moreover, our SA can thoroughly revise out the generic response given by the Seq2Seq, e.g., "*What are you talking about???*" in Examples 3 and 5. The resulting responses are more related to the input, which also explains the improvement of BLEU scores.

## 4 RELATED WORK

Seq2Seq models are a common approach to open-domain dialogue systems. However, Seq2Seq tends to generate short and dull responses. In previous work, Li et al. [9] build a persona-based dialogue system; others explicitly introduce keywords [15] and topics [24] to the generated response. Our work follows the direction of emotional dialogue generation, where we aim to generate a response with an explicitly given emotion [6, 20, 21, 28].

Emotion categorizations have been studied in previous work. For example, Ekman [2] proposes six basic emotions: happiness, sadness, fear, disgust, anger, and surprise. Plutchik [17] proposes a wheel of emotions to model the correlations of 24 emotions. They are widely used for emotion classification [4, 5, 18, 26]. The NLPCC 2017 dataset we used generally follows Ekman's categorization with an additional no-emotion category, except that the emotion "fear" is excluded due to its infrequent occurrences.

In this work, we design a search-based approach for emotional dialogue generation. This follows the recent development for unsupervised text generation, where discrete search is performed towards a heuristically defined objective [7, 10, 12, 14, 19]. By contrast, our search objective is given by a Seq2Seq model trained in a supervised way, but amended by an emotional classifier.

## 5 CONCLUSION

In this study, we propose a search-based emotional dialogue system by simulated annealing. Our method starts with a general response given by standard Seq2Seq. Then, it gradually edits the response, which eventually expresses a specific emotion. We conducted experiments on the NLPCC2017 dataset. Results show that our method outperforms previous state-of-the-art models in terms of the BLEU score and emotion accuracy. Especially, the emotion accuracy is improved by a large extent of 12 percentage points.

## ACKNOWLEDGMENTS
The work is supported in part by the Natural Sciences and Engineering Research Council of Canada (NSERC) under grant Nos. RGPIN2020-04465 and RGPIN-2020-04440. Chenyang Huang is supported by the Borealis AI Graduate Fellowship Program. Lili Mou and Osmar Zaïane are supported by the Amii Fellow Program and the Canada CIFAR AI Chair Program. Lili Mou is supported in part by a donation from DeepMind. This research is also supported by Compute Canada (www.computecanada.ca).